\documentclass[11pt,a4paper]{article}
\usepackage[hyperref]{acl2020}
\usepackage{times}
\usepackage{todonotes}
\usepackage{latexsym}
\usepackage{subcaption}
\usepackage{caption}

\usepackage{microtype}
\usepackage[ruled,vlined]{algorithm2e}

\usepackage{hyperref}       
\usepackage{amsfonts}       
\usepackage{nicefrac}       
\usepackage{lipsum}

\usepackage{subcaption}
\usepackage{booktabs}
\usepackage{verbatim}
\usepackage{amsmath}
\usepackage{graphicx}

\usepackage{footmisc}
\usepackage{float}
\usepackage{placeins}
\newcommand{\squishlist}{ 
   \begin{list}{$\bullet$}
    { \setlength{\itemsep}{0pt}      \setlength{\parsep}{2pt} 
      \setlength{\topsep}{3pt}       \setlength{\partopsep}{0pt}
      \setlength{\leftmargin}{1em} \setlength{\labelwidth}{1em}
      \setlength{\labelsep}{0.5em} } }

\newcommand{\squishend}{
    \end{list}  } 

\newcommand\sysname{XtremeDistil }

\aclfinalcopy 
\def\aclpaperid{2162} 

\makeatletter
\def\blfootnote{\xdef\@thefnmark{}\@footnotetext}
\makeatother

\title{XtremeDistil: Multi-stage Distillation for Massive Multilingual Models}

\author{Subhabrata Mukherjee \\
  Microsoft Research AI \\
  Redmond, WA \\
  \texttt{submukhe@microsoft.com} \\\And
  Ahmed Hassan Awadallah \\
  Microsoft Research AI \\
  Redmond, WA \\
  \texttt{hassanam@microsoft.com} \\}

\begin{document}

\setlength{\abovedisplayskip}{5pt}
\setlength{\belowdisplayskip}{5pt}

\maketitle
\begin{abstract}
Deep and large pre-trained language models are the state-of-the-art for various natural language processing tasks. However, the huge size of these models could be a deterrent to using them in practice. Some recent works use knowledge distillation to compress these huge models into shallow ones. In this work we study knowledge distillation with a focus on multilingual Named Entity Recognition (NER). In particular, we study several distillation strategies and propose a stage-wise optimization scheme leveraging teacher internal representations, that is agnostic of teacher architecture, and show that it outperforms strategies employed in prior works. Additionally, we investigate the role of several factors like the amount of unlabeled data, annotation resources, model architecture and inference latency to name a few. We show that our approach leads to massive compression of teacher models like mBERT by upto $35x$ in terms of parameters and $51x$ in terms of latency for batch inference while retaining $95\%$ of its $F_1$-score for NER over 41 languages.

\end{abstract}

\section{Introduction}

\noindent{\bf Motivation:} Pre-trained language models have shown state-of-the-art performance for various natural language processing applications like text classification, named entity recognition and question-answering. 
%
%
A significant challenge facing practitioners is how to deploy these huge models in practice. For instance, models like BERT Large~\cite{DBLP:conf/naacl/DevlinCLT19}, GPT 2~\cite{radford2019}, Megatron~\cite{Shoeybi2019MegatronLMTM} and T5~\cite{Raffel2019ExploringTL} have $340M$, $1.5B$, $8.3B$ and $11B$ parameters respectively. Although these models are trained offline, during prediction we need to traverse the deep neural network architecture stack involving a large number of parameters. This significantly increases latency and memory requirements.

Knowledge distillation~\cite{DBLP:journals/corr/HintonVD15,DBLP:conf/nips/BaC14} earlier used in computer vision provides one of the techniques to compress huge neural networks into smaller ones. In this,  shallow models (called students) are trained to mimic the output of huge models (called teachers) based on a transfer set. Similar approaches have been recently adopted for language model distillation.


\noindent {\bf Limitations of existing work:} Recent works ~\citep{DBLP:journals/corr/abs-1904-09482,zhu-etal-2019-panlp,DBLP:journals/corr/abs-1903-12136,turc2019wellread} leverage soft logits from teachers as optimization targets for distilling students, with some notable exceptions from concurrent work. \citet{sun2019patient,sanh2019}; \citet{aguilar2019knowledge}; \citet{zhao2019extreme} additionally use internal representations from the teacher as additional signals. 
However, these methods are constrained by architectural considerations like embedding dimension in BERT and transformer architectures. This makes it difficult to massively compress these models (without being able to reduce network width) or adopt alternate architectures. For instance, we observe BiLSTMS as students to be more accurate than Transformers for low latency configurations. Some of the concurrent works~\citep{turc2019wellread}; \citep{zhao2019extreme} adopt pre-training or dual training to distil students of arbitrary architecture. However, pre-training is expensive both in terms of time and computational resources.

Additionally, most of the above works are geared for distilling language models for GLUE tasks~\cite{wang-etal-2018-glue}. There has been some limited exploration of such techniques for sequence tagging tasks like NER~\citep{izsak2019training,Shi_2019} or multilingual tasks~\citep{Tsai_2019}. However, these works also suffer from similar drawbacks as mentioned before.



\noindent {\bf Overview of XtremeDistil:\blfootnote{\hspace{-2em} XtremeDistil: Multilingual pre-\underline{TR}ain\underline{E}d \underline{M}od\underline{E}l Distillation}} 
In this work, we compare distillation strategies used in all the above works and propose a new scheme outperforming prior ones. In this, we leverage teacher internal representations to transfer knowledge to the student. However, in contrast to prior work, we are not restricted by the choice of student architecture. This allows representation transfer from Transformer-based teacher model to BiLSTM-based student model with different embedding dimensions and disparate output spaces. 
We also propose a stage-wise optimization scheme to sequentially transfer most general to task-specific information from teacher to student for better distillation. 


\noindent {\bf Overview of our task:} Unlike prior works mostly focusing on GLUE tasks in a single language, we employ our techniques to study distillation for massive multilingual Named Entity Recognition (NER) over 41 languages. Prior work on multilingual transfer on the same~\cite{rahimi-etal-2019-massively} (MMNER) requires knowledge of source and target language whereby they judiciously select pairs for effective transfer 
resulting in a customized model for each language. In our work, we adopt Multilingual Bidirectional Encoder Representations from Transformer (mBERT) as our teacher and show that it is possible to perform language-agnostic joint NER for all languages with a single model that has a similar performance but massively compressed in contrast to mBERT and MMNER.

Perhaps, the closest work to this work is that of ~\citep{Tsai_2019} where mBERT is leveraged for multilingual NER. We discuss this in details and use their strategy as one of our baselines. We show that our distillation strategy is better leading to a much higher compression and faster inference. We also investigate several unexplored dimensions of distillation like the impact of unlabeled transfer data and annotation resources, choice of multilingual word embeddings, architectural variations and inference latency to name a few. 

Our techniques obtain massive compression of teacher models like mBERT by upto $35x$ in terms of parameters and $51x$ in terms of latency for batch inference while retaining $95\%$ of its performance for massive multilingual NER, and matching or outperforming it for classification tasks.
%
\noindent Overall, our work makes the following {\em contributions}:
\squishlist
    \item {\bf Method:} We propose a distillation method leveraging internal representations and parameter projection that is agnostic of teacher architecture. 
    \item {\bf Inference:} To learn model parameters, we propose stage wise optimization schedule with gradual unfreezing outperforming prior schemes.
    \item {\bf Experiments:} We perform distillation for multilingual NER on 41 languages with massive compression and comparable performance to huge models\footnote{Code and resources available at: \url{https://aka.ms/XtremeDistil}}. We also perform classification experiments on four datasets where our compressed models perform at par with significantly larger teachers.
    \item {\bf Study:} We study the influence of several factors on distillation like the availability of annotation resources for different languages, model architecture, quality of multilingual word embeddings, memory footprint and inference latency.
\squishend

\noindent {\bf Problem Statement:}
Consider a sequence $x=\langle x_k \rangle$ with $K$ tokens and $y=\langle y_k \rangle$ as the corresponding labels. Consider $D_l = \{\langle x_{k,l} \rangle, \langle y_{k,l} \rangle \}$ to be a set of $n$ labeled instances with $X=\{\langle x_{k,l} \rangle \}$ denoting the instances and $Y=\{\langle y_{k,l} \rangle \}$ the corresponding labels. Consider $D_u = \{\langle x_{k,u} \rangle\}$ to be a transfer set of $N$ unlabeled instances from the same domain where 
$n \ll N$. Given a teacher $\mathcal{T}(\theta^t)$, we want to train a student $\mathcal{S}(\theta^s)$ with $\theta$ being trainable parameters such that $|\theta^s| \ll |\theta^t|$ and the student is comparable in performance to the teacher based on some evaluation metric. In the following section, the superscript `t' always represents the teacher and `s' denotes the student.

\section{Related Work}


\noindent{\bf Model compression and knowledge distillation:} Prior works in the vision community dealing with huge architectures like AlexNet and ResNet have addressed this challenge in two ways. Works in model compression use quantization~\citep{DBLP:journals/corr/GongLYB14}, low-precision training and pruning the network, as well as their combination~\citep{HanMao16} to reduce the memory footprint. On the other hand, works in knowledge distillation leverage student teacher models. 
These approaches include using soft logits as targets~\citep{DBLP:conf/nips/BaC14}, increasing the temperature of the softmax to match that of the teacher~\citep{DBLP:journals/corr/HintonVD15} as well as using teacher representations~\citep{DBLP:journals/corr/RomeroBKCGB14} 
(refer to~\citep{DBLP:journals/corr/abs-1710-09282} for a survey). 

\noindent{\bf Recent and concurrent Works:} \citet{DBLP:journals/corr/abs-1904-09482,zhu-etal-2019-panlp,Clark-2019} leverage ensembling to distil knowledge from several multi-task deep neural networks into a single model. 
\citet{sun2019patient,sanh2019};\citet{aguilar2019knowledge} train student models leveraging architectural knowledge of the teacher models which adds architectural constraints (e.g., embedding dimension) on the student. In order to address this shortcoming, more recent works combine task-specific distillation with pre-training the student model with arbitrary embedding dimension but still relying on transformer architectures \citep{turc2019wellread}; \citep{jiao2019tinybert}; \citep{zhao2019extreme}. 

\citet{izsak2019training,Shi_2019} extend these for sequence tagging for Part-of-Speech (POS) tagging and Named Entity Recognition (NER) in English. The one closest to our work \citet{Tsai_2019} extends the above for multilingual NER.

Most of these works rely on general corpora for pre-training and task-specific labeled data for distillation. To harness additional knowledge, \citep{turc2019wellread} leverage task-specific unlabeled data. \citep{DBLP:journals/corr/abs-1903-12136,jiao2019tinybert} use rule-and embedding-based data augmentation in absence of such unlabeled data. 


\section{Models}

\noindent {\bf The Student:}
The input to the model are $E$-dimensional word embeddings for each token. In order to capture sequential information in the sentence, we use a single layer Bidirectional Long Short Term Memory Network (BiLSTM). 
Given a sequence of $K$ tokens, a BiLSTM computes a set of $K$ vectors $h(x_k) = [\overrightarrow{h(x_k)}; \overleftarrow{h(x_k)}]$ as the concatenation of the states generated by a forward $(\overrightarrow{h(x_k)})$ and backward LSTM $(\overleftarrow{h(x_k)})$. Assuming the number of hidden units in the LSTM to be $H$, each hidden state $h(x_k)$ is of dimension $2H$.
%
%
%
Probability distribution for the token label at timestep $k$ is given by:
\begin{equation}
    \label{eq:0}
p^{(s)}(x_k) = softmax(h(x_k) \cdot W^{s})
\end{equation}
\noindent where $W^{s} \in R^{2H.C}$ and $C$ is number of labels. 

Consider one-hot encoding of the token labels, such that $y_{k,l,c}=1$ for $y_{k,l}=c$, and $y_{k,l,c}=0$ otherwise for $c\in C$. The overall cross-entropy loss computed over each token obtaining a specific label in each sequence is given by:
\begin{equation}
\label{eq:1}
    \mathcal{L_{CE}} = -\sum_{x_{l}, y_{l} \in D_l} \sum_k \sum_c y_{k,c, l}\  log\ p_{c}^{(s)}(x_{k,l})
\end{equation}

We train the student model end-to-end minimizing the above cross-entropy loss over labeled data. 

\noindent{\bf The Teacher:}
Pre-trained language models like ELMO~\citep{DBLP:conf/naacl/PetersNIGCLZ18}, BERT~\citep{DBLP:conf/naacl/DevlinCLT19} and GPT~\citep{radford2018improving,radford2019} have shown state-of-the-art performance for several tasks. We adopt BERT as the teacher -- specifically, the multilingual version of BERT (mBERT) with $179MM$ parameters trained over 104 languages with the largest Wikipedias. 
mBERT does not use any markers to distinguish languages during pre-training and learns a single language-agnostic model trained via masked language modeling over Wikipedia articles from all languages. 


\noindent{\bf Tokenization}: Similar to mBERT, we use WordPiece tokenization with $110K$ shared WordPiece vocabulary. 
We preserve casing, remove accents, split on punctuations and whitespace.

\noindent {\bf Fine-tuning the Teacher}: The pre-trained language models are trained for general language modeling objectives. In order to adapt them for the given task, the teacher is fine-tuned end-to-end with task-specific labeled data $D_l$ to learn parameters $\tilde{\theta^t}$ using cross-entropy loss as in Equation~\ref{eq:1}.


\section{Distillation Features}
Fine-tuning the teacher gives us access to its task-specific representations for distilling the student model. To this end, we use different kinds of information from the teacher.

\subsection{Teacher Logits}
Logits as logarithms of predicted probabilities provide a better view of the teacher by emphasizing on the different relationships learned by it across different instances. 
%
Consider $p^t(x_k)$ to be the classification probability of token $x_k$ as generated by the fine-tuned teacher with $logit(p^t(x_k))$ representing the corresponding logits. Our objective is to train a student model with these logits as targets. 
Given the hidden state representation $h(x_k)$ for token $x_k$, we can obtain the corresponding classification score (since targets are logits) as:
\begin{equation}
r^s(x_k) = W^r \cdot h(x_k) + b^r    
\end{equation}
\noindent where $W^r \in R^{C \cdot 2H}$ and $b^r \in R^C$ are trainable parameters and $C$ is the number of classes. We want to train the student neural network end-to-end by 
minimizing the element-wise mean-squared error between the classification scores given by the student and the target logits from the teacher as:

{\small
\begin{multline}
\label{eq:4}
    \mathcal{L_{LL}} = \frac{1}{2}\sum_{x_u \in D_u} \sum_k ||r^s(x_{k,u})-logit(p^t(x_{k,u};\tilde{\theta_t}))||^2
\end{multline}
\vspace{-2em}
}
%
%
%
\subsection{Internal Teacher Representations}

\noindent {\bf Hidden representations:} 
Recent works~\citep{sun2019patient,DBLP:journals/corr/RomeroBKCGB14} have shown the hidden state information from the teacher to be helpful as a hint-based guidance for the student. Given a large collection of task-specific unlabeled data, we can transfer the teacher's knowledge to the student via its hidden representations. However, this poses a challenge in our setting as the teacher and student models have different architectures with disparate output spaces.


Consider $h^s(x_k)$ and $z_l^t(x_k; \tilde{\theta_t})$ to be the representations generated by the student and the $l^{th}$ deep layer of the fine-tuned teacher respectively for a token $x_k$. Consider $x_u \in D_u$ to be the set of unlabeled instances. We will later discuss the choice of the teacher layer $l$ and its impact on distillation.

\noindent {\bf Projection:} To make all output spaces compatible, we perform a non-linear projection of the parameters in student representation $h^s$ to have same shape as teacher representation $z_l^t$ for each token $x_k$: 
\begin{equation}
 \tilde{z}^s(x_k)=Gelu(W^f \cdot h^s(x_k) + b^f)
\end{equation}
\noindent where $W^f \in R^{|z_l^t| \cdot 2H}$ is the projection matrix, $b^f \in R^{|z_l^t|}$ is the bias, and {\em Gelu} (Gaussian Error Linear Unit)~\citep{DBLP:journals/corr/HendrycksG16} is the non-linear projection function. $|z_l^t|$ represents the embedding dimension of the teacher. This transformation aligns the output spaces of the student and teacher and allows us to accommodate arbitrary student architecture. Also note that the projections (and therefore the parameters) are shared across tokens at different timepoints.

The projection parameters are learned by minimizing the $KL$-divergence (KLD) between the student and the $l^{th}$ layer teacher representations:
\begin{multline}
\label{eq:6}
    \mathcal{L_{RL}} = \sum_{x_u \in D_u} \sum_k KLD( \tilde{z}^s(x_{k,u}),z_l^t(x_{k,u};\tilde{\theta_t})) 
\end{multline}

\vspace{-0.5em}
\noindent {\bf Multilingual word embeddings:} A large number of parameters reside in the word embeddings. For mBERT a shared multilingual WordPiece vocabulary of $V=110K$ tokens and embedding dimension of $D=768$ leads to $92MM$ parameters. To have massive compression, we cannot directly incorporate mBERT embeddings in our model. Since we use the same WordPiece vocabulary, we are likely to benefit more from these embeddings than from Glove~\citep{DBLP:conf/emnlp/PenningtonSM14} or FastText~\citep{bojanowski2016enriching}.

We use a dimensionality reduction algorithm like Singular Value Decomposition (SVD) to project the mBERT word embeddings to a lower dimensional space. Given mBERT word embedding matrix of dimension $V$$\times$$D$, 
SVD finds the best $E$-dimensional representation that minimizes sum of squares of the projections (of rows) to the subspace. 



\section{Training} 
\vspace{-0.5em}

We want to optimize the loss functions for {\em representation} $\mathcal{L_{RL}}$, {\em logits} $\mathcal{L_{LL}}$ and {\em cross-entropy} $\mathcal{L_{CE}}$. These optimizations can be scheduled differently to obtain different training regimens as follows.

\subsection{Joint Optimization}

In this, we optimize the following losses jointly:

{\small
\begin{multline}
    \frac{1}{|D_l|} \sum_{\{x_l, y_l\}\in D_l} \alpha \cdot \mathcal{L_{CE}}(x_l, y_l) +\\  \frac{1}{|D_u|} \sum_{\{x_u, y_u\}\in D_u} \bigg( \beta \cdot \mathcal{L_{RL}} (x_u, y_u) + \gamma \cdot \mathcal{L_{LL}} (x_u, y_u) \bigg)
\label{eq:71}
\end{multline}
}

\noindent where $\alpha, \beta$ and $\gamma$ weigh the contribution of different losses. A high value of $\alpha$ makes the student focus more on easy targets; whereas a high value of $\gamma$ leads focus to the difficult ones. 
The above loss is computed over two different task-specific data segments. The first part involves cross-entropy loss over labeled data, whereas the second part involves representation and logit loss over unlabeled data. 


\setlength{\textfloatsep}{0.1cm}
\setlength{\floatsep}{0.1cm}
\begin{algorithm}[t]
\SetAlgoLined
\small
Fine-tune teacher on $D_l$ and update $\tilde{\theta^t}$ \;
\For {stage in \{1,2,3\}} {
Freeze all student layers $l' \in \{1 \cdots L\}$\;
\If{stage=1} {
$output$ = $\tilde{z}^s(x_u)$ \;
$target$ = teacher representations on $D_u$ from the $l^{th}$ layer as $z_l^t(x_u;\tilde{\theta^t})$ \;
$loss$ = $\mathcal{R_{RL}}$ \;
}
\If{stage=2} {
$output$ = $r^s(x_u)$ \;
$target$ = teacher logits on $D_u$ as $logit(p^t(x_u; \tilde{\theta^t}))$  \;
$loss$ = $\mathcal{R_{LL}}$ \;
}
\If{stage=3} {
$output$ = $p^s(x_l)$ \;
$target$ = $y_l \in D_l$ \;
$loss$ = $\mathcal{R_{CE}}$ \;
}
 \For{ layer $l' \in \{L \cdots 1\}$} {
    Unfreeze $l'$ \;
    Update parameters $\theta^s_{l'},\theta^s_{l'+1}\cdots\theta^s_{L}$ by minimizing the optimization $loss$ between student $output$ and teacher $target$
    }
    }
 \caption{Multi-stage distillation.}
 \label{algo:1}
\end{algorithm}

\subsection{Stage-wise Training}

Instead of optimizing all loss functions jointly, we propose a stage-wise scheme to gradually transfer most general to task-specific representations from teacher to student. In this, we first train the student to mimic teacher representations from its $l^{th}$ layer by optimizing $\mathcal{R_{RL}}$ on {unlabeled data}. The student learns the parameters for word embeddings ($\theta^w$), BiLSTM ($\theta^b$) and projections $\langle W^f,b^f \rangle$.

In the second stage, we optimize for the cross-entropy $\mathcal{R_{CE}}$ and logit loss $\mathcal{R_{LL}}$ jointly on both labeled and unlabeled data respectively to learn the corresponding parameters $W^s$ and $\langle W^r, b^r \rangle$.

The above can be further broken down in two stages, where we sequentially optimize logit loss $\mathcal{R_{LL}}$ on unlabeled data and then optimize cross-entropy loss $\mathcal{R_{CE}}$ on labeled data. 
Every stage learns parameters conditioned on those learned in previous stage followed by end-to-end fine-tuning. 

\subsection{Gradual Unfreezing}

One potential drawback of end-to-end fine-tuning for stage-wise optimization is `catastrophic forgetting'~\citep{DBLP:conf/acl/RuderH18} where the model forgets information learned in earlier stages. To address this, we adopt gradual unfreezing -- where we tune the model one layer at a time starting from the configuration at the end of previous stage. 

We start from the top layer that contains the most task-specific information and allow the model to configure the task-specific layer first while others remain frozen. The latter layers are gradually unfrozen one by one and the model trained till convergence. Once a layer is unfrozen, it maintains the state. When the last layer (word embeddings) is unfrozen, the entire network is trained end-to-end. The order of this unfreezing scheme (top-to-bottom) is reverse of that in \citep{DBLP:conf/acl/RuderH18} and we find this to work better in our setting with the following intuition. At the end of the first stage on optimizing $\mathcal{R_{RL}}$, the student learns to generate representations similar to that of the $l^{th}$ layer of the teacher. Now, we need to add only a few task-specific parameters ($\langle W^r, b^r \rangle$) to optimize for logit loss $\mathcal{R_{LL}}$ with all others frozen. Next, we {\em gradually} give the student more flexibility to optimize for task-specific loss by tuning the layers below where the number of parameters increases with depth ($|\langle W^r, b^r \rangle| \ll |\theta_b| \ll |\theta_w|$).

We tune each layer for $n$ epochs and restore model to the best configuration based on validation loss on a held-out set. Therefore, the model retains best possible performance from any iteration. Algorithm~\ref{algo:1} shows overall processing scheme.

\begin{table}
    \centering
    \small
    \begin{tabular}{lcccc}
        \toprule
         { Dataset} & { Labels} & { Train} & { Test} & { Unlabeled} \\\midrule
         {\em NER} &&&&\\
         Wikiann-41 & 11 & 705K & 329K & 7.2MM \\\midrule
         {\em Classification} &&&&\\
         IMDB & 2 & 25K & 25K & 50K \\
         DBPedia & 14 & 560K & 70K & - \\
         AG News & 4 & 120K & 7.6K & - \\
         Elec & 2 & 25K & 25K & 200K \\
         \bottomrule
    \end{tabular}
    \vspace{-1em}
    \caption{Full dataset summary.}
    \label{tab:dataset}
\end{table}

\section{Experiments}

\begin{table}
    \centering
    \small
    \begin{tabular}{p{4.5cm}rrr}
    \toprule
Work	& PT & TA & Distil.\\
	\midrule
\citet{sanh2019} & Y & Y & D1\\
\citet{turc2019wellread} & Y & N & D1\\\midrule
\citet{DBLP:journals/corr/abs-1904-09482,zhu-etal-2019-panlp,Shi_2019,Tsai_2019,DBLP:journals/corr/abs-1903-12136,izsak2019training,Clark-2019} & N & N & D1\\\midrule
\citet{sun2019patient} & N & Y & D2 \\
\citet{jiao2019tinybert} & N & N & D2 \\
\citet{zhao2019extreme} & Y & N & D2 \\\midrule
\sysname (ours) & N & N & D4\\
\bottomrule
    \end{tabular}
    \vspace{-1em}
    \caption{Different distillation strategies. D1 leverages soft logits with hard labels. D2 uses representation loss. PT denotes pre-training with language modeling. TA depicts students constrained by teacher architecture.} 
    \label{tab:strategy}
\end{table}

\begin{table*}[t]
    \centering
    \small
    \begin{tabular}{lcccc}
        \toprule
Strategy & Features & 	Transfer = 0.7MM & Transfer = 1.4MM & Transfer = 7.2MM \\\midrule
D0 & Labels per lang. &	71.26 (6.2) &	- &	-  \\\midrule
D0-S & Labels across all lang. &	81.44 (5.3) &	- &	-\\\midrule
D1 & Labels and Logits  &	82.74 (5.1) &	84.52 (4.8) &	85.94 (4.8)\\
D2 & Labels, Logits and Repr. &	82.38 (5.2) &	83.78 (4.9) &	85.87 (4.9)\\\midrule
D3.1 & (S1) Repr. (S2) Labels and Logits & 83.10 (5.0) &	84.38 (5.1) &	86.35 (4.9)\\
D3.2 & + Gradual unfreezing &	86.77 (4.3) &	87.79 (4.0) &	88.26 (4.3)\\\midrule
D4.1 & (S1) Repr. (S2) Logits (S3) Labels &	84.82 (4.7)	& 87.07 (4.2) &	87.87 (4.1)\\
D4.2 & + Gradual unfreezing & 	87.10 (4.2) &	{\bf 88.64 (3.8)}	& 88.52 (4.1)\\
\bottomrule
    \end{tabular}
    \vspace{-1em}
    \caption{Comparison of several strategies with average $F_1$-score (and standard deviation) across 41 languages over different transfer data size. $S_i$ depicts separate stages and corresponding optimized loss functions. }
    \vspace{-1.5em}
    \label{tab:strategy-comparison}
\end{table*}

\noindent{\bf Dataset Description:} 
We evaluate our model \sysname for multilingual NER on 41 languages and the same setting as in~\citep{rahimi-etal-2019-massively}. This data has been derived from the WikiAnn NER corpus~\citep{pan-etal-2017-cross} and partitioned into training, development and test sets. All the NER results are reported in this test set for a fair comparison between existing works. We report both the average $F_1$-score ($\mu$) and standard deviation $\sigma$ between scores across 41 languages for phrase-level evaluation. Refer to Figure~\ref{fig:comparison} for languages codes and distribution of training labels across languages. We also perform experiments with data from four other domains (refer to Table~\ref{tab:dataset}): IMDB~\citep{DBLP:conf/acl/MaasDPHNP11}, SST-2~\cite{socher-etal-2013-parsing} and Elec~\citep{DBLP:conf/recsys/McAuleyL13} for sentiment analysis for movie and electronics product reviews, DbPedia~\citep{DBLP:conf/nips/ZhangZL15} and Ag News~\citep{DBLP:conf/nips/ZhangZL15} for topic classification of Wikipedia and news articles. 

\noindent{\bf NER Tags:} The NER corpus uses IOB2 tagging strategy with entities like LOC, ORG and PER. Following mBERT, we do not use language markers and share these tags across all languages. We use additional syntactic markers like \{CLS, SEP, PAD\} and `X' for marking segmented wordpieces contributing a total of 11 tags (with shared `O'). 


\subsection{Evaluating Distillation Strategies}
\noindent{\bf Baselines:} A trivial baseline (D0) is to learn models {\em one per language} using only corresponding labels for learning. This can be improved by merging all instances and sharing information across all languages (D0-S). Most of the concurrent and recent works (refer to Table~\ref{tab:strategy} for an overview) leverage logits as optimization targets for distillation (D1). A few exceptions also use teacher internal representations along with soft logits (D2). 
For our model we consider multi-stage distillation, where we first optimize representation loss followed by jointly optimizing logit and cross-entropy loss (D3.1) and further improving it by gradual unfreezing of neural network layers (D3.2). Finally, we optimize the loss functions sequentially in three stages (D4.1) and improve it further by unfreezing mechanism (D4.2). We further compare all strategies while varying the amount of unlabeled transfer data for distillation (hyper-parameter settings in Appendix).

\noindent{\bf Results:} From Table~\ref{tab:strategy-comparison}, we observe all strategies that share information across languages to work better (D0-S vs. D0) with the soft logits adding more value than hard targets (D1 vs. D0-S). Interestingly, we observe simply combining representation loss with logits (D3.1 vs. D2) hurts the model. We observe this strategy to be vulnerable to the hyper-parameters ($\alpha, \beta, \gamma$ in Eqn.~\ref{eq:71}) used to combine multiple loss functions. We vary hyper-parameters in multiples of 10 and report best numbers.

Stage-wise optimizations remove these hyper-parameters and improve performance. We also observe the gradual unfreezing scheme to improve both stage-wise distillation strategies significantly.

Focusing on the data dimension, we observe all models to improve as more and more unlabeled data is used for transferring teacher knowledge to student. However, we also observe the improvement to slow down after a point where additional unlabeled data does not yield significant benefits. Table~\ref{tab:improvement} shows the gradual performance improvement in \sysname after every stage and unfreezing various neural network layers.

\subsection{Performance, Compression and Speedup}
\begin{table}
        \small
    \centering
    \begin{tabular}{lccc}
        \toprule
Stage & Unfreezing Layer & $F_1$ & Std. Dev.\\\bottomrule
2 & Linear ($\langle W^r, b^r \rangle$) &	0 & 0\\
2 & Projection  ($\langle W^f, b^f \rangle$) &	2.85 & 3.9\\
2 & BiLSTM ($\theta_b$) &	81.64 & 5.2\\
2 & Word Emb ($\theta_w$) &	85.99 & 4.4\\\midrule
3 & Softmax ($W^s$) &	86.38 & 4.2\\
3 & Projection ($\langle W^f, b^f \rangle$) &	87.65 & 3.9\\
3 & BiLSTM ($\theta_b$) &	88.08 & 3.9\\
3 & Word Emb ($\theta_w$) & 	88.64 & 3.8\\
\bottomrule
    \end{tabular}
    \vspace{-0.5em}
    \caption{Gradual $F_1$-score improvement over multiple distillation stages in \sysname.}
    \label{tab:improvement}
\end{table}

\begin{figure}
\centering
\begin{subfigure}{.5\textwidth}
  \centering
    \includegraphics[width=\linewidth]{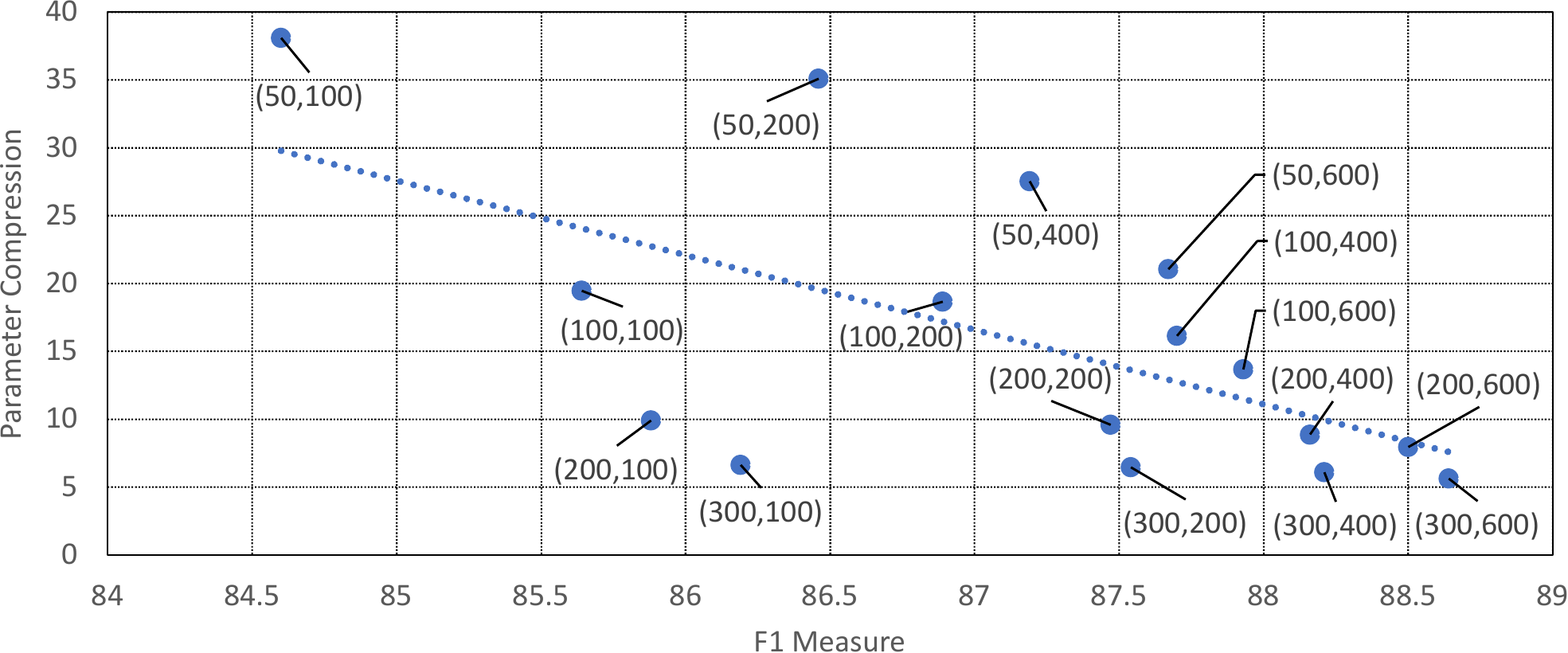}
    \vspace{-1.5em}
  \caption{Parameter compression vs. $F_1$-score.}
  \label{fig:param-comparison}
\end{subfigure}

\begin{subfigure}{.5\textwidth}
  \centering
  \includegraphics[width=\linewidth]{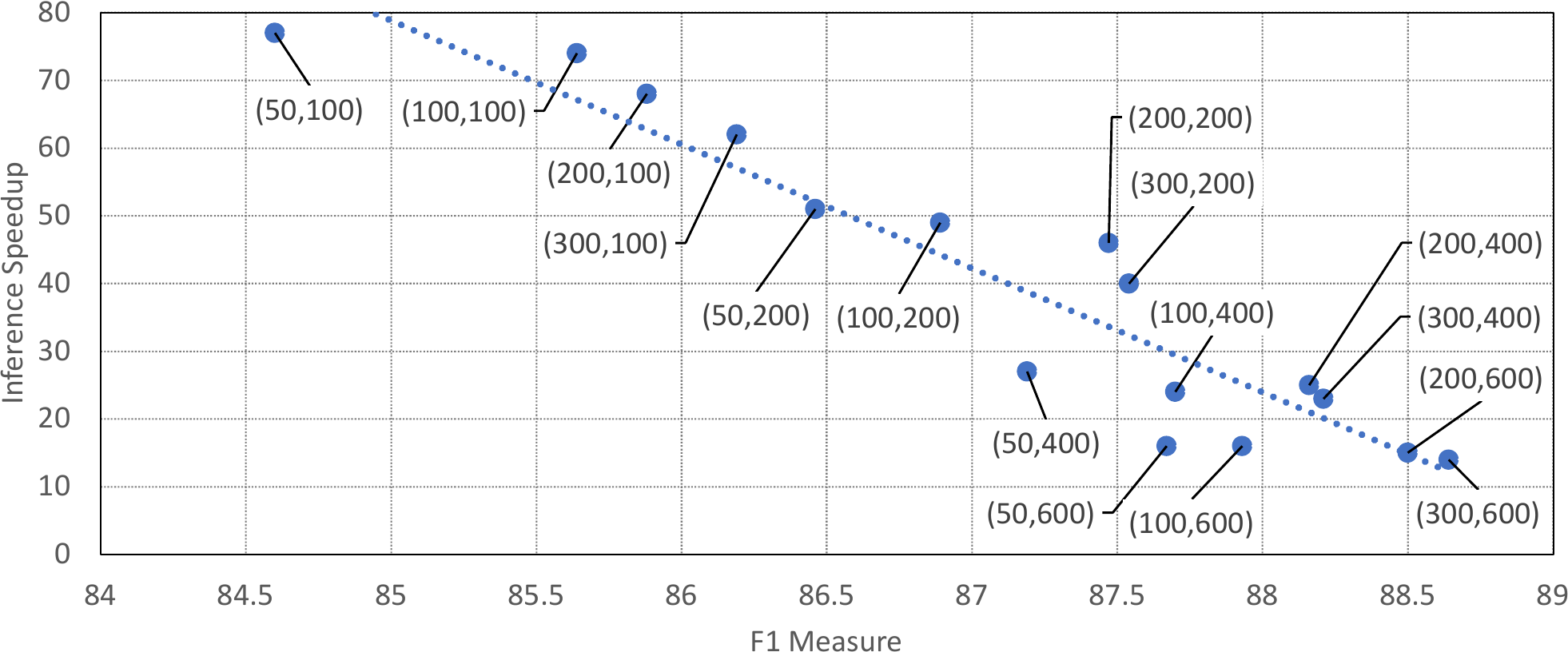}
  \caption{Inference speedup vs. $F_1$-score.}
  \label{fig:latency-comparison}
\end{subfigure}
\vspace{-0.5em}
\caption{Variation in \sysname $F_1$-score with parameter and latency compression against mBERT. Each point in the linked scatter plots represents a configuration with corresponding embedding dimension and BiLSTM hidden states as ($E,H$). Data point ($50, 200$) in both figures correspond to $35x$ compression and $51x$ latency speedup.}
\label{fig:compression}
\end{figure}

\begin{table}
    \centering
    \small
    \begin{tabular}{lcc}
        \toprule
        Model & Avg. $F_1$ & Std. Dev\\\midrule
        mBERT-single~\citep{DBLP:conf/naacl/DevlinCLT19} & 90.76 & 3.1\\
        mBERT~\cite{DBLP:conf/naacl/DevlinCLT19} &	91.86 & 2.7\\
MMNER~\citep{rahimi-etal-2019-massively} & 89.20 & 2.8\\
\sysname (ours) &	88.64 & 3.8\\
\bottomrule
    \end{tabular}
    \vspace{-0.5em}
    \caption{$F_1$-score comparison of different models with standard deviation across 41 languages.}
    \label{tab:f-score-comparison}
\end{table}

\begin{figure*}
\centering
  \includegraphics[width=\linewidth]{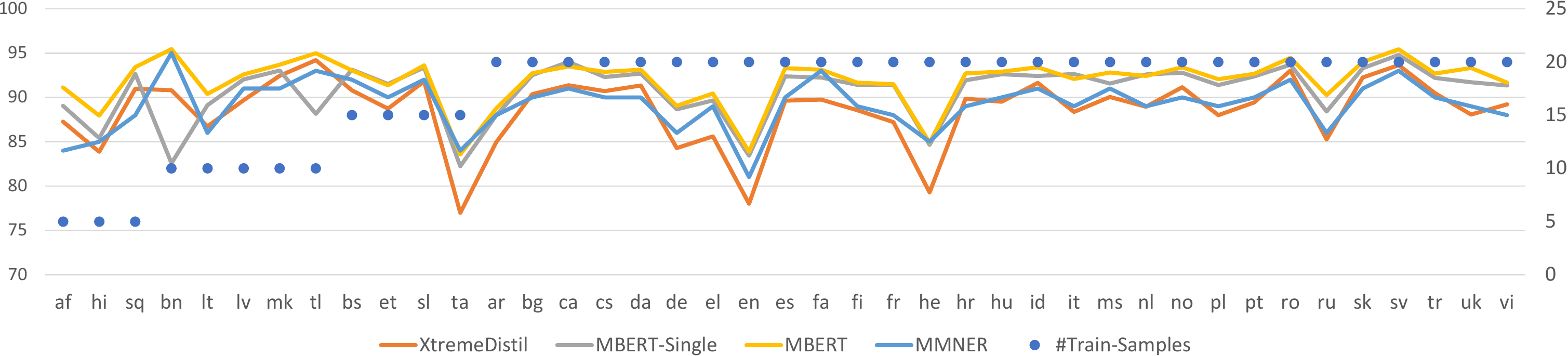}
    \vspace*{-1em}
  \caption{$F_1$-score comparison for different models across 41 languages. The y-axis on the left shows the scores, whereas the axis on the right (plotted against blue dots) shows the number of training labels (in thousands).}
  \label{fig:comparison}
  \vspace{-1em}
\end{figure*}

\noindent{\bf Performance:} We observe \sysname in Table~\ref{tab:f-score-comparison} to perform competitively with other models. mBERT-single models are fine-tuned per language with corresponding labels, whereas mBERT is fine-tuned with data across all languages. MMNER results are reported from ~\citet{rahimi-etal-2019-massively}.

Figure~\ref{fig:comparison} shows the variation in $F_1$-score across different languages with variable amount of training data for different models.  We observe all the models to follow the general trend with some aberrations for languages with less training labels.

\noindent{\bf Parameter compression:} \sysname performs at par with MMNER in terms of $F_1$-score while obtaining at least $41x$ compression. Given $L$ languages, MMNER learns ($L-1$) ensembled and distilled models, one for each target language. Each of the MMNER language-specific models is comparable in size to our single multilingual model. We learn a single model for all languages, thereby, obtaining a compression factor of at least $L=41$.


Figure~\ref{fig:param-comparison} shows the variation in $F_1$-scores of \sysname and compression against mBERT with different configurations corresponding to the embedding dimension ($E$) and number of BiLSTM hidden states ($2\times H$). We observe that reducing the embedding dimension leads to great compression with minimal performance loss. Whereas, reducing the BiLSTM hidden states impacts the performance more and contributes less to the compression. 

\noindent{\bf Inference speedup:} We compare the runtime inference efficiency of mBERT and our model in a single P100 GPU for batch inference (batch size = 32) on $1000$ queries of sequence length $32$. We average the time taken for predicting labels for all the queries for each model aggregated over $100$ runs. Compared to batch inference, the speedups are less for online inference (batch size = 1) at $17x$ on Intel(R) Xeon(R) CPU (E5-2690 v4 @2.60GHz) (refer to Appendix for details).

Figure~\ref{fig:latency-comparison} shows the variation in $F_1$-scores of \sysname and inference speedup against mBERT with different (linked) parameter configurations as before. As expected, the performance degrades with gradual speedup. We observe that parameter compression does not necessarily lead to an inference speedup. Reduction in the word embedding dimension leads to massive model compression, however, it does not have a similar effect on the latency. The BiLSTM hidden states, on the other hand, constitute the real latency bottleneck. One of the best configurations leads to $35x$ compression, $51x$ speedup over mBERT retaining nearly $95\%$ of its performance.  

\begin{table}
    \centering
    \small
    \begin{tabular}{lcc}
        \toprule
Model & \#Transfer Samples & $F_1$\\\midrule
MMNER & - & 62.1 \\\midrule
mBERT &	- & 79.54\\\midrule
\sysname & 4.1K & 19.12\\
& 705K &	76.97	\\
& 1.3MM &	77.17	\\
& 7.2MM &	77.26\\
\bottomrule	
  \end{tabular}
    \vspace{-0.5em}
    \caption{$F_1$-score comparison for low-resource setting with $100$ labeled samples per language and transfer set of different sizes for \sysname.}
    \label{tab:few-shot}
\end{table}

\vspace{-0.5em}
\subsection{Low-resource NER and Distillation}

Models in all prior experiments are trained on $705K$ labeled instances across all languages. In this setting, we consider only $100$ labeled samples for each language with a total of $4.1K$ instances. From Table~\ref{tab:few-shot}, we observe mBERT to outperform MMNER by more than $17$ percentage points with \sysname closely following suit. 

Furthermore, we observe our model's performance to improve with the transfer set size depicting the importance of unlabeled transfer data for knowledge distillation. As before, a lot of additional data has marginal contribution. 

\vspace{-0.5em}
\subsection{Word Embeddings}

From Table~\ref{tab:word-emb}, we observe that random initialization of word embeddings works quite well. Multilingual $300d$ FastText embeddings~\cite{bojanowski2016enriching} leads to minor improvement due to $38\%$ overlap between FastText tokens and mBERT wordpieces. English $300d$ Glove does much better. We experiment with dimensionality reduction techniques and find SVD to work better. Surprisingly, it leads to marginal improvement over mBERT embeddings before reduction. As expected, mBERT embeddings after fine-tuning perform better than that from pre-trained checkpoints.

\begin{table}[h]
    \centering
    \small
    \begin{tabular}{p{5cm}p{0.5cm}p{0.5cm}}
        \toprule
Word Embedding & $F_1$-score & Std. Dev.\\\midrule
SVD + mBERT (fine-tuned) &		{\bf 88.64} &	{\bf 3.8}\\
mBERT (fine-tuned)	&	88.60	& 3.9\\
SVD + mBERT (pre-trained) &	88.54 & 3.9\\
PCA + PPA (d=14)~\cite{Raunak_2019} &		88.35 &	3.9\\
PCA + PPA (d=17)~\cite{Raunak_2019}	&	88.25	& 4.0\\
Glove~\cite{DBLP:conf/emnlp/PenningtonSM14}	&	88.16	 & 4.0\\
FastText~\cite{bojanowski2016enriching} &		87.91 & 3.9\\
Random &		87.43	&  4.1\\
\bottomrule
  \end{tabular}
    \vspace{-0.5em}
    \caption{Impact of using various word embeddings for initialization on multilingual distillation. SVD, PCA, and Glove uses $300$-dimensional word embeddings.}
    \label{tab:word-emb}
\end{table}

\subsection{Architectural Considerations}

\noindent {\bf Which teacher layer to distil from?} The topmost teacher layer captures more task-specific knowledge. However, it may be difficult for a shallow student to capture this knowledge given its limited capacity. On the other hand, the less-deep representations at the middle of teacher model are easier to mimic by shallow student. From Table~\ref{tab:layer} we observe the student to benefit most from distilling the $6^{th}$ or $7^{th}$ layer of the teacher.

\begin{table}[h]
    \centering
    \small
    \begin{tabular}{lcc}
        \toprule
Layer ($l$) & $F_1$-score & Std. Dev.\\\midrule
11 &	88.46 & 3.8\\
9 & 88.31 & 3.8\\
{\bf 7} &	{\bf 88.64} & {\bf 3.8}\\
6 &	88.64 & 3.8\\
4 &	88.19 & 4\\
2 &	88.50 & 4\\
1 &	88.51 & 4\\
\bottomrule	
  \end{tabular}
    \vspace{-0.5em}
    \caption{Comparison of \sysname performance on distilling representations from $l^{th}$ mBERT layer.}
    \label{tab:layer}
\end{table}
\begin{figure}
\centering
  \includegraphics[width=\linewidth]{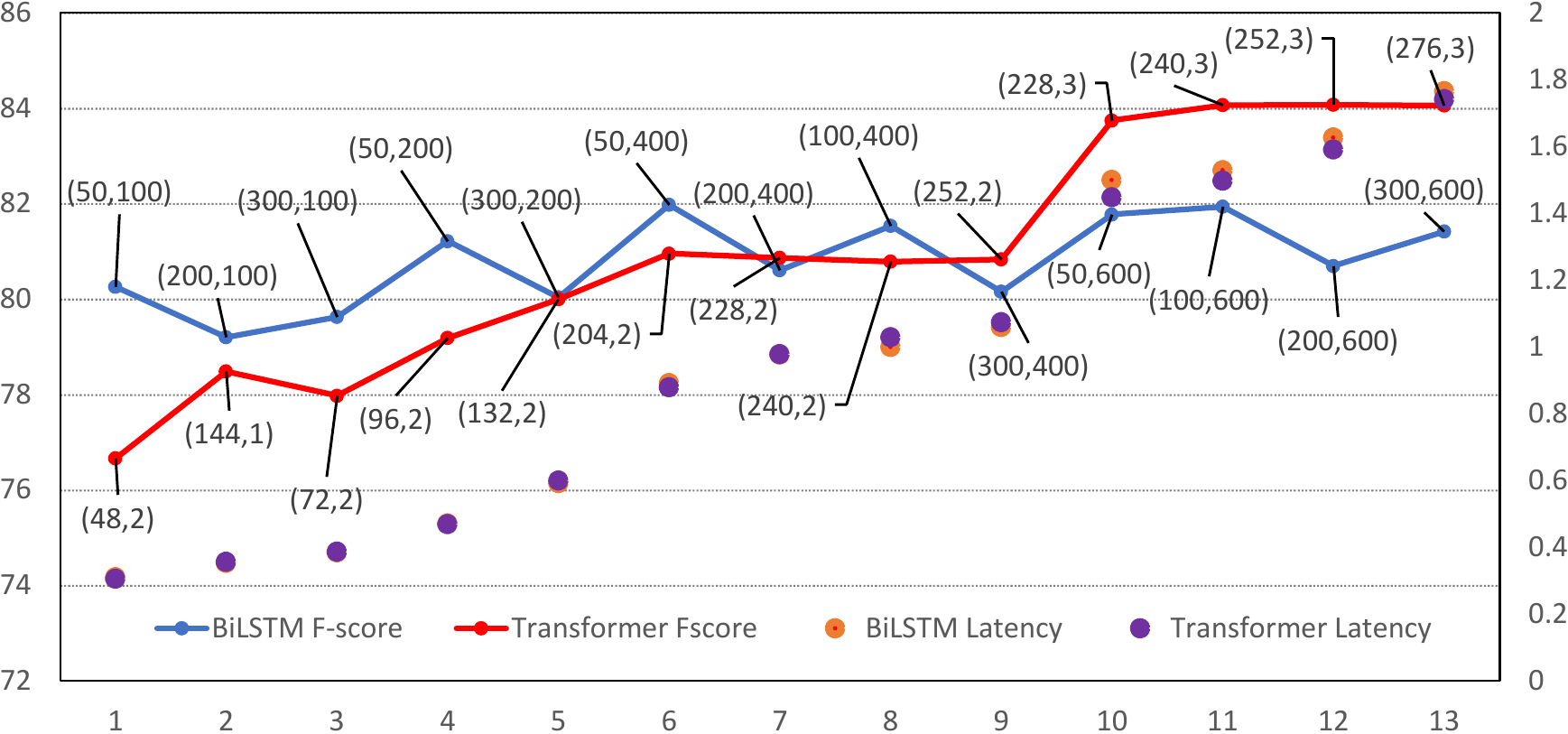}
  \vspace{-1.5em}
  \caption{BiLSTM and Transformer $F_1$-score (left y-axis) vs. inference latency (right y-axis) in 13 different settings with corresponding embedding dimension and width / depth of the student as $(E, W/D)$.}
  \label{fig:arch}
\end{figure}

\noindent {\bf Which student architecture to use for distillation?} Recent works in distillation leverage both BiLSTM and Transformer as students. 
In this experiment, 
we vary the embedding dimension and hidden states for BiLSTM-, and embedding dimension and depth for Transformer-based students to obtain configurations with similar inference latency. Each of $13$ configurations in Figure~\ref{fig:arch} depict $F_1$-scores obtained by students of different architecture but similar latency (refer to Table~\ref{tab:arch-comparison} in Appendix for statistics) -- for strategy D0-S in Table~\ref{tab:strategy-comparison}. We observe that for low-latency configurations BiLSTMs with hidden states $\{2$$\times$$100,2$$\times$$200\}$ work better than $2$-layer Transformers. Whereas, the latter starts performing better with more than $3$-layers although with a higher latency compared to the aforementioned BiLSTM configurations. 

\subsection{Distillation for Text Classification}

We switch gear and focus on classification tasks. In contrast to sequence tagging, we use the last hidden state of the BiLSTM as the final sentence representation for projection, regression and softmax.

Table~\ref{tab:teacher} shows the distillation performance of \sysname with different teachers on four benchmark text classification datasets. We observe the student to almost match the teacher performance for all of the datasets. The performance also improves with a better teacher, although the improvement is marginal as the student capacity saturates. 

\begin{table}
    \centering
    \small
    \begin{tabular}{lccccc}
    \toprule
{ Dataset} & Student &{ Distil} & { Distil } &  { BERT} & {	BERT}\\
 & {\scriptsize no distil.} & {\scriptsize (Base)} & {\scriptsize (Large)} & {\scriptsize Base} & {\scriptsize Large} \\
\midrule
Ag News & 89.71 &	92.33 &	94.33 & 92.12	& {\bf 94.63}\\
IMDB &	89.37 & 91.22 &	91.70 &	91.70 &	{\bf 93.22}\\
Elec & 90.62 & 93.55 &	93.56 &	93.46 &	{\bf 94.27}\\
DbPedia	& 98.64 & 99.10	& 99.06 &	{\bf 99.26}	& 99.20\\
\bottomrule
    \end{tabular}
    \vspace{-1em}
    \caption{Distillation performance with BERT.}
    \label{tab:teacher}
\end{table}

Table~\ref{tab:less} shows the distillation performance with only $500$ labeled samples per class. The distilled student improves over the non-distilled version by $19.4$ percent and matches the teacher performance for all of the tasks demonstrating the impact of distillation for low-resource settings.

\begin{table}[]
    \centering
    \small
    \begin{tabular}{lccc}
\toprule
{ Dataset} & { Student} & { Student}  &	{ BERT}\\
 &  { no distil.} & { with distil.}  &	{ Large}\\
\midrule
AG News & 85.85 &	{\bf 90.45} &	90.36\\
IMDB & 61.53 &	89.08 &	{\bf 89.11}\\
Elec & 65.68	& {\bf 91.00}	& 90.41\\
DBpedia & 96.30 &	{\bf 98.94}	& {\bf 98.94}\\
\bottomrule
    \end{tabular}
    \vspace{-1em}
    \caption{Distillation with BERT Large on $500$ labeled samples per class.}
    \label{tab:less}
\end{table}

\begin{table}
    \centering
    \small
    \begin{tabular}{lcc}
        \toprule
Model & Transfer Set & Acc.\\\midrule
BERT Large Teacher & - & 94.95\\
\sysname &  SST+Imdb & 93.35\\\midrule
BERT Base Teacher &	- & 92.78\\
\sysname &  SST+Imdb & 92.89\\
\citet{sun2019patient} & SST & 92.70\\
\citet{turc2019wellread} & SST+IMDB & 91.10\\
\bottomrule
  \end{tabular}
    \vspace{-.5em}
    \caption{Model accuracy on of SST-2 (dev. set).}
    \label{tab:sst}
\end{table}

\noindent {\bf Comparison with other distillation techniques:} SST-2~\cite{socher-etal-2013-parsing} from GLUE~\cite{wang-etal-2018-glue} has been used as a test bed for other distillation techniques for single instance classification tasks (as in this work). Table~\ref{tab:sst} shows the accuracy comparison of such methods reported in SST-2 development set with the same teacher. 

We extract $11.7MM$ sentences from all IMDB movie reviews in Table~\ref{tab:dataset} to form the unlabeled transfer set for distillation. We obtain the best performance on distilling with BERT Large (uncased, whole word masking model) than BERT Base -- demonstrating a better student performance with a better teacher and outperforming other methods.

\section{Summary}

{\noindent \bf Teacher hidden representation and distillation schedule:} Internal teacher representations help in distillation, although a naive combination hurts the student model. We show that a distillation schedule with stagewise optimization, gradual unfreezing with a cosine learning rate scheduler (D4.1 + D4.2 in Table~\ref{tab:strategy-comparison}) obtains the best performance. We also show that the middle layers of the teacher are easier to distil by shallow students and result in the best performance (Table~\ref{tab:layer}). Additionally, the student performance improves with bigger and better teachers (Tables~\ref{tab:teacher} and~\ref{tab:sst}).

\noindent {\bf Student architecture:} We compare different student architectures like BiLSTM and Transformer in terms of configuration and performance (Figure~\ref{fig:arch}, Table~\ref{tab:arch-comparison} in Appendix), and observe BiLSTM to perform better at low-latency configurations, whereas the Transformer outperforms the former with more depth and higher latency budget. 

\noindent {\bf Unlabeled transfer data:} We explored the data dimension in Tables~\ref{tab:strategy-comparison} and~\ref{tab:few-shot} and observed unlabeled data to be the key for knowledge transfer from deep pre-trained teachers to shallow students and bridge the performance gap. 

We observed a moderate amount of unlabeled transfer samples ($0.7$ - $1.5$ million) lead to the best student, whereas larger amounts of transfer data does not result in significant gains. This is particularly helpful for low-resource NER (with only $100$ labeled samples per language as in Table~\ref{tab:few-shot}). 

\noindent {\bf Performance trade-off:} Parameter compression does not necessarily reduce inference latency, and vice versa. We explore model performance in terms of parameter compression, inference latency and $F_1$ to show the trade-off for distillation in Figure~\ref{fig:compression} and Table~\ref{tab:compression-config} in Appendix. 

\noindent {\bf Multilingual word embeddings:} Random initialization of word embeddings work well. A better initialization, which is also parameter-efficient, is given by Singular Value Decomposition (SVD) over fine-tuned mBERT word embeddings with the best performance for downstream task (Table~\ref{tab:word-emb}). 

\noindent {\bf Generalization:} The outlined distillation techniques and strategies are model-, architecture-, and language-agnostic and can be easily extended to arbitrary tasks and languages, although we only focus on NER and classification in this work. 

\noindent {\bf Massive compression:} Our techniques demonstrate massive compression ($35x$ for parameters) and inference speedup ($51x$ for latency) while retaining $95\%$ of the teacher performance allowing deep pre-trained models to be deployed in practice. 


\section{Conclusions}

We develop \sysname for massive multi-lingual NER and classification that performs close to huge pre-trained models like MBERT but with massive compression and inference speedup. Our distillation strategy leveraging teacher representations agnostic of its architecture and stage-wise optimization schedule outperforms existing ones. We perform extensive study of several distillation dimensions like the impact of unlabeled transfer set, embeddings and student architectures, and make interesting observations outlined in summary.



\bibliography{acl2020}
\bibliographystyle{acl_natbib}

\newpage

\appendix

\section{Appendices}
\label{sec:appendix}

\subsection{Implementation}
The model uses Tensorflow backend. Code and resources available at: \url{https://aka.ms/XtremeDistil}.

\subsection{Parameter Configurations}
All the analyses in the paper --- {\em except} compression and speedup experiments that vary embedding dimension $E$ and BiLSTM hidden states $H$ --- are done with the following model configuration in Table~\ref{tab:config} with the best $F_1$-score. Optimizer Adam is used with cosine learning rate scheduler ($lr\_high=0.001, lr\_low=1e-8$). 

The model corresponding to the $35x$ parameter compression and $51x$ speedup for batch inference uses $E=50$ and $H=2\times200$.

\begin{table}[h]
    \centering
    \small
    \begin{tabular}{ll}
        \toprule
Parameter & Value\\\midrule
SVD + MBERT word emb. dim. & $E=300$\\
BiLSTM hidden states & $H=2\times 600$\\
Dropout & 0.2\\
Batch size & 512\\
Teacher layer & 7\\
Optimizer & Adam \\
\bottomrule	
  \end{tabular}
    \caption{\sysname config. with best $F_1=88.64$.}
    \label{tab:config}
\end{table}

Following hyper-parameter tuning was done to select dropout rate and batch size at the start of the parameter tuning process.

\begin{table}[h]
  \centering
  \small
    \begin{tabular}{rr}
    \toprule
    Dropout Rate & $F_1$-score\\\midrule
    1e-4 & 87.94 \\
    0.1   & 88.36 \\
    {\bf 0.2}   & {\bf 88.49} \\
    0.3   & 88.46 \\
    0.6   & 87.26 \\
    0.8   & 85.49 \\\bottomrule
    \end{tabular}%
      \caption{Impact of dropout.}
  \label{tab:addlabel}%
\end{table}%

\begin{table}[h]
  \centering
  \small
    \begin{tabular}{rr}
        \toprule
    Batch size & $F_1$-score\\\midrule
    128   & 87.96 \\
    {\bf 512} & {\bf 88.4} \\
    1024  & 88.24 \\
    2048  & 88.13 \\
    4096  & 87.63 \\\bottomrule
    \end{tabular}%
      \caption{Impact of batch size.}
  \label{tab:addlabel}%
\end{table}%

\begin{table*}[t]
  \centering
  \small
    \begin{tabular}{rrrrrrr|rrrrr}
    \toprule
    \multicolumn{7}{c|}{BiLSTM}                            & \multicolumn{5}{c}{Transformer} \\\midrule
    \multicolumn{1}{l}{Emb} & \multicolumn{1}{l}{Hidden} &       & \multicolumn{1}{l}{F1} &       & \multicolumn{1}{l}{Params (MM)} & \multicolumn{1}{l|}{Latency} & \multicolumn{1}{l}{Emb} & \multicolumn{1}{l}{Depth} & \multicolumn{1}{l}{Params (MM)} & \multicolumn{1}{l}{Latency} & \multicolumn{1}{l}{F1} \\\midrule
    50    & 100   &       & 80.26 &       & 4.7   & 0.311 & 48    & 2     & 4.4   & 0.307 & 76.67 \\
    200   & 100   &       & 79.21 &       & 18.1  & 0.354 & 144   & 1     & 13.4  & 0.357 & 78.49 \\
    300   & 100   &       & 79.63 &       & 27    & 0.385 & 72    & 2     & 6.7   & 0.388 & 77.98 \\
    50    & 200   &       & 81.22 &       & 5.1   & 0.472 & 96    & 2     & 9     & 0.47  & 79.19 \\
    300   & 200   &       & 80.04 &       & 27.7  & 0.593 & 132   & 2     & 12.5  & 0.6   & 80 \\
    50    & 400   &       & 81.98 &       & 6.5   & 0.892 & 204   & 2     & 19.7  & 0.88  & 80.96 \\
    200   & 400   &       & 80.61 &       & 20.2  & 0.978 & 228   & 2     & 22.1  & 0.979 & 80.87 \\
    100   & 400   &       & 81.54 &       & 11.1  & 1     & 240   & 2     & 23.3  & 1.03  & 80.79 \\
    300   & 400   &       & 80.16 &       & 29.4  & 1.06  & 252   & 2     & 24.6  & 1.075 & 80.84 \\
    50    & 600   &       & 81.78 &       & 8.5   & 1.5   & 228   & 3     & 22.7  & 1.448 & 83.75 \\
    100   & 600   &       & 81.94 &       & 13.1  & 1.53  & 240   & 3     & 24    & 1.498 & 84.07 \\
    200   & 600   &       & 80.7  &       & 22.5  & 1.628 & 252   & 3     & 25.3  & 1.591 & 84.08 \\
    300   & 600   &       & 81.42 &       & 31.8  & 1.766 & 276   & 3     & 28    & 1.742 & 84.06 \\\bottomrule
    \end{tabular}%
    \vspace{-1em}
  \caption{Pairwise BiLSTM and Transformer configurations (with varying embedding dimension, hidden states and depth) vs. latency and $F_1$ scores for distillation strategy $D0-S$.}
  \label{tab:arch-comparison}%
\end{table*}%

\begin{table*}[h]
  \centering
  \small
    \begin{tabular}{cccccccc}
    \toprule
    \multicolumn{1}{l}{Embedding} & \multicolumn{1}{l}{BiLSTM} & \multicolumn{1}{l}{$F_1$-score} & \multicolumn{1}{l}{Std. Dev.} & \multicolumn{1}{l}{Params (MM)} & \multicolumn{1}{l}{Params(Compression)} & \multicolumn{1}{l}{Speedup (bsz=32)} & \multicolumn{1}{l}{Speedup (bsz=1)} \\
    \midrule
    300   & 600   & 88.64 & 3.8   & 31.8  & 5.6   & 14    & 8 \\
    200   & 600   & 88.5  & 3.8   & 22.5  & 8     & 15    & 9 \\
    300   & 400   & 88.21 & 4     & 29.4  & 6.1   & 23    & 11 \\
    200   & 400   & 88.16 & 3.9   & 20.2  & 8.9   & 25    & 12 \\
    100   & 600   & 87.93 & 4.1   & 13.1  & 13.7  & 16    & 9 \\
    100   & 400   & 87.7  & 4     & 11.1  & 16.1  & 24    & 13 \\
    50    & 600   & 87.67 & 4     & 8.5   & 21.1  & 16    & 10 \\
    300   & 200   & 87.54 & 4.1   & 27.7  & 6.5   & 40    & 15 \\
    200   & 200   & 87.47 & 4.2   & 18.7  & 9.6   & 46    & 16 \\
    50    & 400   & 87.19 & 4.3   & 6.5   & 27.5  & 27    & 13 \\
    100   & 200   & 86.89 & 4.2   & 9.6   & 18.6  & 49    & 15 \\
    50    & 200   & 86.46 & 4.3   & 5.1   & 35.1  & 51    & 16 \\
    300   & 100   & 86.19 & 4.3   & 27    & 6.6   & 62    & 16 \\
    200   & 100   & 85.88 & 4.4   & 18.1  & 9.9   & 68    & 17 \\
    100   & 100   & 85.64 & 4.5   & 9.2   & 19.5  & 74    & 15 \\
    50    & 100   & 84.6  & 4.7   & 4.7   & 38.1  & 77    & 16 \\\bottomrule
    \end{tabular}%
    \vspace{-1em}
  \caption{Parameter compression and inference speedup vs. $F_1$-score with varying embedding dimension and BiLSTM hidden states. Online inference is in Intel(
R) Xeon(R) CPU (E5-2690 v4 @2.60GHz) and batch inference is in a single P100 GPU for distillation strategy $D4$.}
  \label{tab:compression-config}%
\end{table*}%

\begin{table*}[h]
  \centering
  \small
    \begin{tabular}[t]{lrrrrr}
    \toprule
     Lang  & \#Train & Ours & BERT & MBERT & MMNER \\\midrule
    af    & 5     & 87    & 89    & 91    & 84 \\
    hi    & 5     & 84    & 85    & 88    & 85 \\
    sq    & 5     & 91    & 93    & 93    & 88 \\
    bn    & 10    & 91    & 83    & 95    & 95 \\
    lt    & 10    & 87    & 89    & 90    & 86 \\
    lv    & 10    & 90    & 92    & 93    & 91 \\
    mk    & 10    & 92    & 93    & 94    & 91 \\
    tl    & 10    & 94    & 88    & 95    & 93 \\
    bs    & 15    & 91    & 93    & 93    & 92 \\
    et    & 15    & 89    & 92    & 91    & 90 \\
    sl    & 15    & 92    & 93    & 94    & 92 \\
    ta    & 15    & 77    & 82    & 84    & 84 \\
    ar    & 20    & 85    & 88    & 89    & 88 \\
    bg    & 20    & 90    & 93    & 93    & 90 \\
    ca    & 20    & 91    & 94    & 93    & 91 \\
    cs    & 20    & 91    & 92    & 93    & 90 \\
    da    & 20    & 91    & 93    & 93    & 90 \\
    de    & 20    & 84    & 89    & 89    & 86 \\
    el    & 20    & 86    & 90    & 90    & 89 \\
    en    & 20    & 78    & 83    & 84    & 81 \\
    es    & 20    & 90    & 92    & 93    & 90 \\\bottomrule
    \end{tabular}
    \begin{tabular}[t]{lrrrrr}
    \toprule
    Lang  & \#Train & Ours & BERT & MBERT & MMNER \\\midrule
    fa    & 20    & 90    & 92    & 93    & 93 \\
    fi    & 20    & 89    & 91    & 92    & 89 \\
    fr    & 20    & 87    & 91    & 91    & 88 \\
    he    & 20    & 79    & 85    & 85    & 85 \\
    hr    & 20    & 90    & 92    & 93    & 89 \\
    hu    & 20    & 90    & 93    & 93    & 90 \\
    id    & 20    & 92    & 92    & 93    & 91 \\
    it    & 20    & 88    & 93    & 92    & 89 \\
    ms    & 20    & 90    & 92    & 93    & 91 \\
    nl    & 20    & 89    & 93    & 92    & 89 \\
    no    & 20    & 91    & 93    & 93    & 90 \\
    pl    & 20    & 88    & 91    & 92    & 89 \\
    pt    & 20    & 89    & 92    & 93    & 90 \\
    ro    & 20    & 93    & 94    & 94    & 92 \\
    ru    & 20    & 85    & 88    & 90    & 86 \\
    sk    & 20    & 92    & 93    & 94    & 91 \\
    sv    & 20    & 94    & 95    & 95    & 93 \\
    tr    & 20    & 90    & 92    & 93    & 90 \\
    uk    & 20    & 88    & 92    & 93    & 89 \\
    vi    & 20    & 89    & 91    & 92    & 88 \\
    \\\bottomrule
    \end{tabular}
  \caption{$F_1$-scores of different models per language. BERT represents MBERT fine-tuned separately for each language. Other models including \sysname (ours) is jointly fine-tuned over all languages.}
  \label{tab:addlabel}%
\end{table*}%






\end{document}